\RequirePackage{etoolbox}
\patchcmd{\bibliographystyle}{#1}{ieeetr}{}{}
\documentclass{acta_hungarica}

\usepackage{graphics} 
\usepackage{algorithm}
\usepackage{algorithmic}
\usepackage{multirow}
\usepackage{url}
\usepackage{epsfig} 
\usepackage{bm}
\usepackage{mathptmx} 
\usepackage{balance}
\usepackage{amsmath} 
\usepackage{amssymb}  
\usepackage{caption}
\usepackage{subcaption}

\actaVolume{14}
\actaNumber{12}
\actaYear{2019}

\begin{document}

\title{Raven: Open Surgical Robotic Platforms}
\shorttitle{Raven Review and Prospect}

\author[1]{Yangming Li}
\author[2]{Blake Hannaford}
\author[3]{Jacob Rosen}
\affil[1]{Department of Electrical Computer and Telecommunications Engineering Technology, Rochester Institute of Technology, Rochester, NY, USA 14623, yangming.li@rit.edu}
\affil[2]{Departments of Electrical Engineering, University of Washington, Seattle, WA, USA 98195, blake@uw.edu}
\affil[3]{Department of Mechanical and Aerospace Engineering, University of California, Los Angeles, CA, USA  90095, jacobrosen@ucla.edu}
\shortauthor{Y. Li}

\maketitle

\begin{abstract}
The Raven I and the Raven II surgical robots, as open research platforms, have been serving the robotic surgery research community for ten years. The paper 1) briefly presents the Raven I and the Raven II robots, 2) reviews the recent publications that are built upon the Raven robots, aim to be applied to the Raven robots, or are directly compared with the Raven robots, and 3) uses the Raven robots as a case study to discuss the popular research problems in the research community and the trend of robotic surgery study. Instead of being a thorough literature review, this work only reviews the works formally published in the past three years and uses these recent publications to analyze the research interests, the popular open research problems, and opportunities in the topic of robotic surgery.
\end{abstract} 


\section{Introduction}
\label{sec:intro}

In the past two decades, research on surgical robotics has made impressive progress. Surgical robots successfully transferred from research labs to operating rooms and even became the standard care for a number of surgeries. 

Because of the high value of surgical tasks, the robotic surgery research has focused on the development and the validation of the devices and the techniques that are clinically practical from the very beginning. On one hand, realistically validating the robotic system promotes surgical robotics research, on the other hand, such validation raises the difficulty of surgical robot research as developing surgical robots often requires a significant amount of resources and technical accumulation. 

The Raven I\cite{ravenI} and II\cite{ravenII} surgical robotic platforms were created to serve the robotic surgery research community as open platforms, designed for the study of robotic assisted Minimally Invasive Surgeries (MISs). The Raven robots are also compact in size, and low cost, making them widely adopted by 18 research institutes worldwide.

The Raven I and II surgical robots are designed to provide both hardware (mechanics and electronics) and software to support open research innovations. The Raven software is fully open-source\cite{ravengit} and is made ROS compatible, in order to support research in both the high-level robotic functional development and the low-level position and velocity control. In order to further promote the open robotic surgery research, Raven simulator and the online Raven access interface are under development.

This paper will assess the impact of Raven I and II by surveying the Raven related literature published in the past three years. Through reviewing the Raven literature, we attempt to summarize some of the recent progress in the research field and identify unmet needs and challenges.



This rest of this paper is organized as follows: Section II briefly describes the Raven I and II platforms. Section III quantifies Raven citations and analyzes the trends. Section IV analyzes and summarizes the Raven citations according to the research topics and contributions. Based on the analysis, discussions are made in the last section to conclude this work.

\section{Surgical Robotics and Raven I and II Platforms}
\subsection{Overview}
\begin{figure}[h!]
  \centering
  \includegraphics[width=0.95\textwidth]{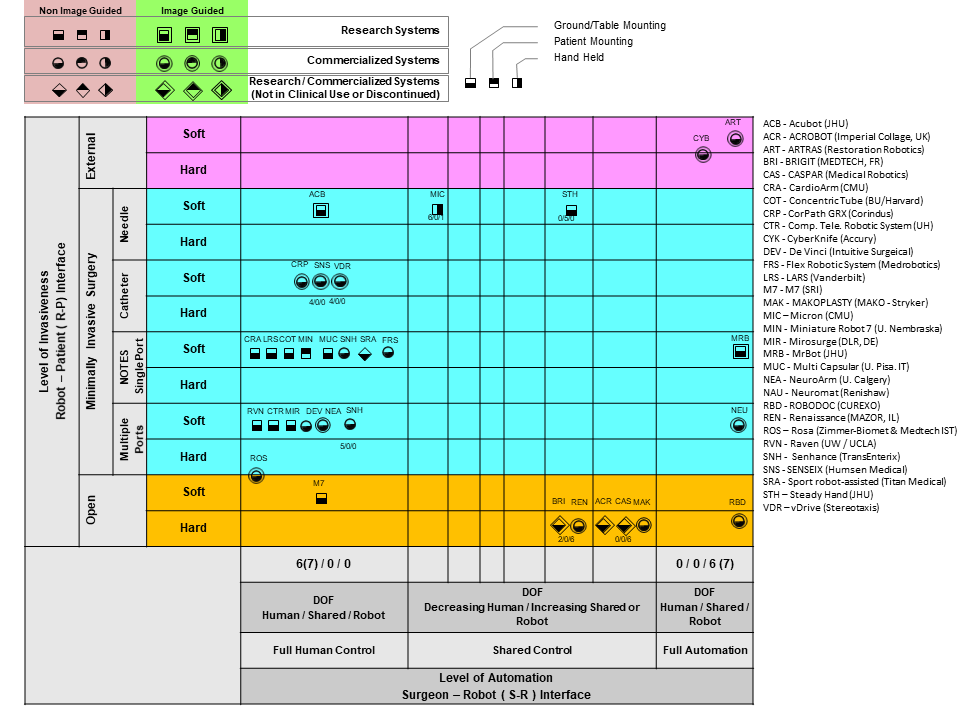}
  \caption{Classification of 2018 surgical robotic systems based on a Surgeon-Robot (S-R) interface (horizontal axis) defining the level of automation and a Robot-Patient (R-P) interface dictating the level of invasiveness.}
  \label{fig:surgicalRobots}
\end{figure}

The introduction of surgical robotics into the operating room offers a significant breakthrough to potentially improve the quality and outcome of surgery. In robotic surgeries, two human-machine interfaces are established: the surgeon-robot interface (S-R) and the patient-robot interface (R-P). These two interfaces may be used to classify the various surgical robotic systems described as of the end of 2018 (Fig. \ref{fig:surgicalRobots}). From the figure we can also see the development of surgical robotics and the shift of research interests. The detailed introduction to the surgical robots listed in Fig. \ref{fig:surgicalRobots} can be found in \cite{rosen2011surgical}.

\begin{figure}[h!]
   \centering
   \begin{subfigure}[t]{0.45\textwidth}
     \centering
     \includegraphics[height=4cm]{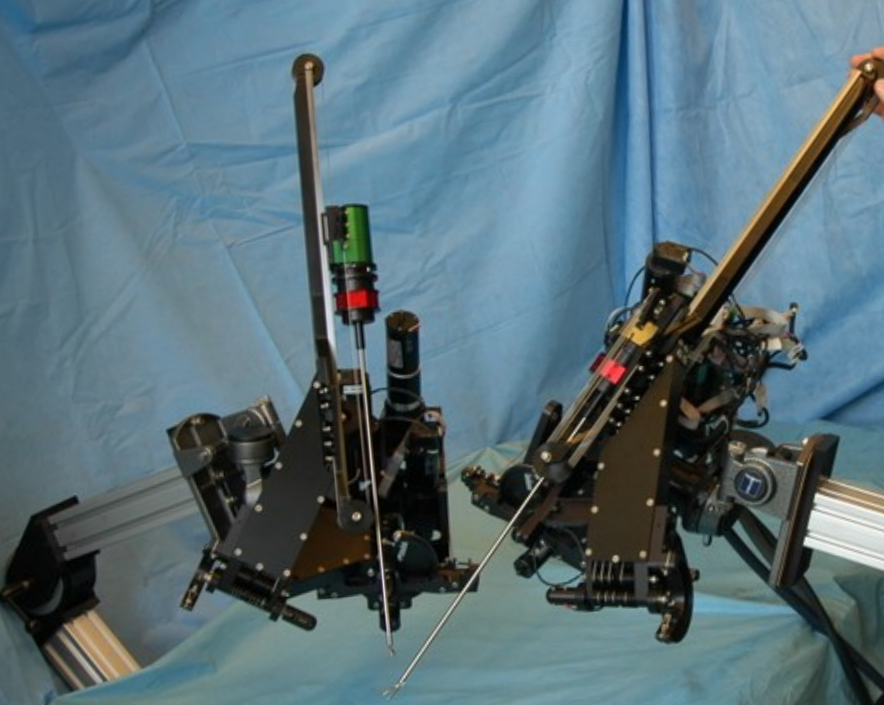}
        \caption{Raven I}
    \end{subfigure}%
    ~ 
    \begin{subfigure}[t]{0.55\textwidth}
      \centering
      \includegraphics[height=4cm]{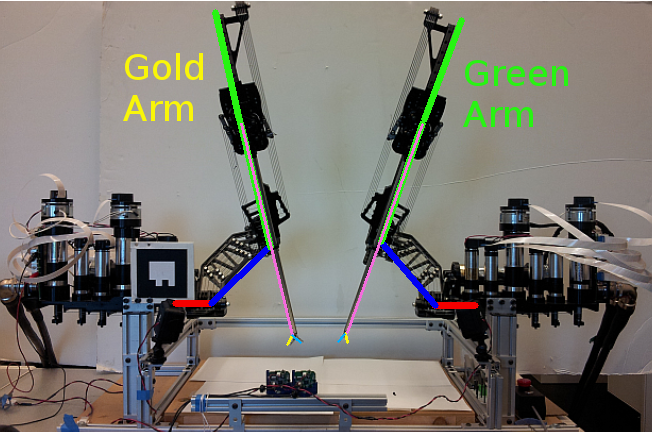}
        \caption{Raven II}
      \end{subfigure}
    \caption{The Raven I and II Surgical Robotic Platforms.}
    \label{fig:raven}
\end{figure}

The Raven I and II surgical robots were first presented as full surgical platforms in \cite{ravenI} and \cite{ravenII} respectively (Figure~\ref{fig:raven}). The Raven II evolved from the Raven I with improved mechanical design and software. This section summarizes some important features they have in common. From the hardware perspective, the Raven robots provide seven Degrees of Freedom(DoF) in manual Minimally Invasive Surgeries (MISs) (x, y, and z positions, three axes of rotation, and grasper open/close), through a seven DoF cable-driven robotic manipulator. The platform was specifically designed and optimized for MIS, as the remote center was built in with a spherical mechanism, and neither physical constraints nor control algorithms are needed to prevent tangential motions and forces, which could injure the patient at the insertion site in the abdominal wall. From the software perspective, the Raven software is developed based on real-time Linux, augmented by a Programmable Logic Controller (PLC) based safety mechanism. The entire raven source code is open-source, which includes kinematics/dynamics based control and teleoperation.

\subsection{Raven Hardware}
The Raven robot hardware consists of the master console (the surgeon site) and the slave robotic arms (the patient site).

The master console consists of devices that teleoperate the slave robots' movements, and a foot pedal that couples/decouples the master/slave motion synchronization. The Raven I robots used PHANTOM Omni devices to control the motion, and the Raven II robots can work with various control devices.

The slave robot arms and instruments contain the core of Raven mechanical design effort. Each Raven arm contains one rotational shoulder joint, one rotational elbow joint, one transnational insertion/retraction link, two grasper tip (finger) rotational links (one for rotation, the other one for open/close grasping), and two rotational wrist links (for the two different wrist rotational motions). The Raven II and the Raven I share the same fundamental mechanical design, such as having built-in remote centers, and 7 DoFs. The differences between the two platforms are: 1) the Raven II has the more compact mechanical design, 2) the Raven II improves the tool interface design.

The detailed hardware designs, including the mechanics, the DH parameters, the kinematics, and the electronics can be found in \cite{ravenI} and \cite{ravenII}.

\subsection{Raven Software}

The Raven software design started from the safety requirements. The control system was built upon real-time Linux, and works at 1000 Hz. In order to achieve the software system reliability, the Raven software contains four states: initialization, pedal up, pedal down and an emergency stop. The software failures in the first three states are constantly monitored with a watchdog timer and a separate hardware PLC guarantees the failures are reliably caught and the system immediately switches to the emergency stop once failures are caught.

Both the Raven I and the Raven II software contain modules for hardware control and monitoring, forward and inverse kinematics, gravity compensation, and closed-loop control. The main differences between the two are: 1) the Raven I software is based on RTAI and the Raven II uses RT-Linux, 2) the Raven II software provides the ROS compatibility and contains more modules such as dynamics, state estimation, interactive force estimation, and autonomous motion planning. The latest Raven software can be found: https://github.com/uw-biorobotics/raven2.

In order to facilitate the robotic surgery research, an ongoing project is underway to unify the programming environment between the Raven platforms and another prominent research platform the da Vinci Research Kit\cite{dvrk} through 1) open APIs, 2) remote access, 3) simulators. The latest open-sourced APIs and simulators can be found here: https://github.com/collaborative-robotics.

\section{Raven: An Open Platform for Robotic Surgery Study}
The Raven platforms attract robotic surgery researchers through its open source software stack and flexible hardware interfaces. This section studies selected publications which cite the Raven I\cite{ravenI} and the Raven II\cite{ravenII} introduction papers. Only the publications in the last 3 years (Jan. 2016~Oct. 2018) are reviewed. According to Google Scholar (scholar.google.com), the two Raven papers\cite{ravenI,ravenII} were cited 197 times since Jan. 2016. Only \emph{research papers} that are formally published are included, which leads to 69 publications. 

\begin{figure}[h!]
  \centering
  \includegraphics[width=0.65\textwidth]{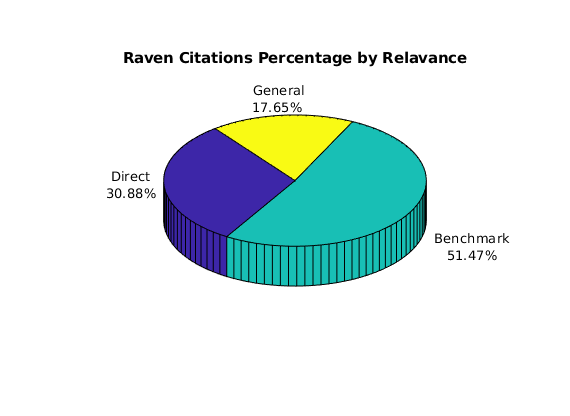}
  \caption{Percentage of Raven Citations by Relevance. The Raven citations are divided into three groups: 1) the ones that directly used Raven (indicated by `'Direct''), 2)  the ones not used Raven but related to robotic surgeries (referred as ``Benchmark''), and 3) the ones not used Raven and out of the scope of robotic surgeries (referrred as ``General'').}
  \label{fig:totalbyrelavance}
\end{figure}

According to the relevance, the Raven citations are categorized into two groups: 1) the research described does not use the Raven platforms, 2) the research describes ones use the Raven platforms in their researches, we refer the later as ``Direct''. In the former category, we further divide these publications into two groups, the ones related to robotic surgeries (referred as ``Benchmark'') and the ones out of the scope of robotic surgeries (referred as ``General''). The percentages of the three categories are shown in Fig. \ref{fig:totalbyrelavance}. The figure shows that it is reasonable to use the Raven platforms as a sample to study the prospect of robotic surgery study because it shows the Raven platforms have a broader impact on robotic surgery research and general robotic research.

\begin{figure}[h!]
  \centering
  \includegraphics[width=0.65\textwidth]{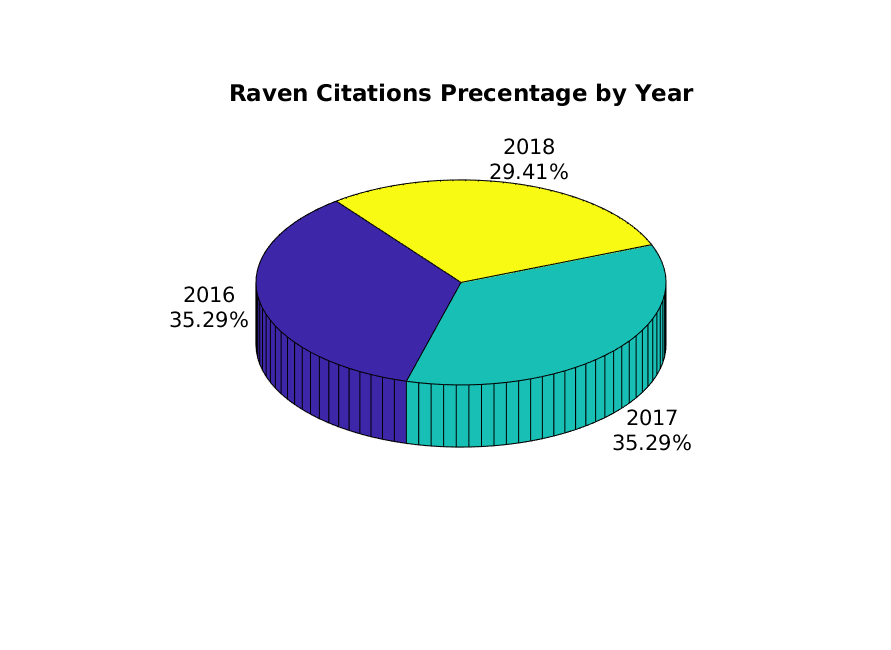}
  \caption{Raven Citations Distribution over Years. The figure shows the total number of the formally published Raven citations in 2016, 2017 and 2018. From the figure we can see that the Raven related research is stable in the number of publications.}
  \label{fig:totalbyyear}
\end{figure}

\begin{figure}[h!]
  \centering
  \includegraphics[width=0.65\textwidth]{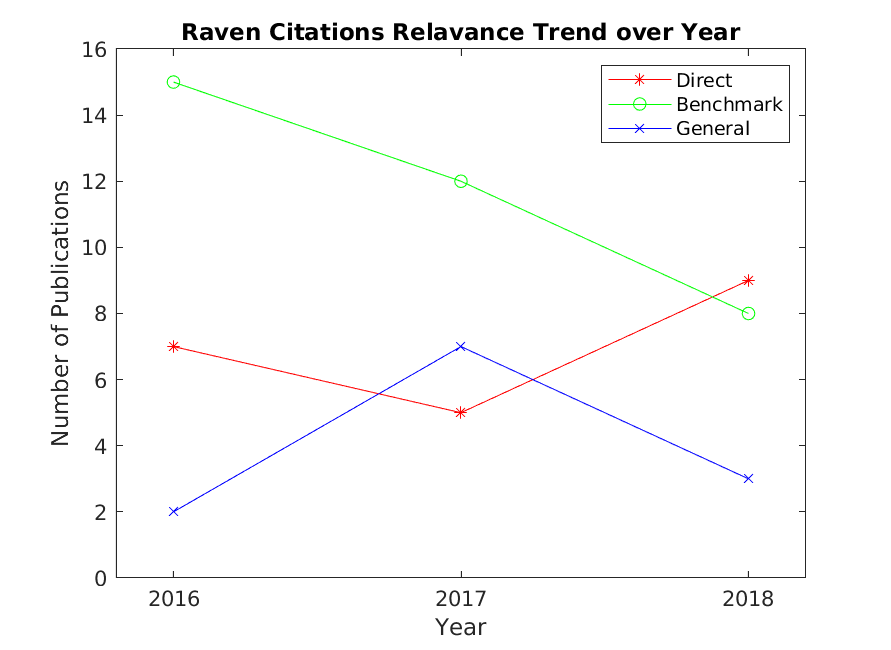}
  \caption{Raven Citations Relevance by Years. It categories the Raven citations by relevance and shows the numbers of citations in the three categories changed over years. From the figure we can see the total number of direct Raven research papers climbs up and matches the number of the publications used the Raven robots as a benchmark.}
  \label{fig:relavancebyyear}
\end{figure}

We also summarize the Raven citations over the past three years to show the popularity of Raven related researches, as shown in Fig.\ref{fig:totalbyyear}. The categorized publications based on the relevance to the Raven platforms is also compared with respect to time, as shown in Fig.\ref{fig:relavancebyyear}, to show the popularity and the trend of the related research. 
%
%
%

It is interesting to see how the publications in each of the relevant categories vary with time (Fig. \ref{fig:relavancebyyear}). From the figure we can see that the quantities of the surgical related Raven citations (the green line indicates the citations in the category ``Benchmark'', the red line indicates the citations in the category ``Direct'', the blue line indicates the citations in the category ``General'', we can see that the Raven Platforms have bigger impact in the robotic surgery research community than in the general robotic research community, and the publications that use Raven .

\subsection{Research Citing Raven but Not Using Raven}

As the Raven platforms are widely used in the robotic surgery research, many works have cited the Raven system but not used it in the research they describe.  It may be instructive to analyze these papers to the extent they portray directions and frontiers of surgical robotics research today. 

Several groups citing the Raven system have developed complete or nearly complete \emph{surgical robotic systems}\cite{kim2017development,ai2017design,reichenbach2017telesurgery,liu2017development,zhou2017development,kim2016development,marinho2016using}. These systems often address new surgical procedures (such as pediatric cases or needle guidance)\cite{kim2017development,liu2017development,zhou2017development} novel delivery modes\cite{reichenbach2017telesurgery} or integration of industrial manipulators into surgery\cite{marinho2016using}

Numerous groups have developed \emph{new hardware} influenced by the Raven design 
\cite{niu2018improved,diodato2018soft,niu2018dimensional,xu2018kinematic,bazman2018dexterous, nisar2017design,kim2017s,hwang2017single,rosen2017roboscope,marinho2016using,cubrich2016four}
or focused on numerical optimization of mechansims\cite{niu2017multi}. Key aspects of these designs are minimally invasive character, often making a contribution such as novel mechanisms for decoupling motion at the laparoscopic entry port\cite{niu2018improved}, decoupling drive axes\cite{xu2018kinematic}, or reducing weight and size\cite{kim2017s}.

Novel mechanical design directions include integration with soft robotics\cite{diodato2018soft}, 
elbowed instrument design\cite{hwang2017single}, 
mechanical decoupling design\cite{niu2018improved}, and
mechanical integration of novel force/torque sensors\cite{kim2017s}.

Other works focus on kinematic issues related to surgical robotics \cite{torabi2018manipulability,singh2018modular,marinho2016using,yang2016kinematic}, which are often approached from the point of view of improved motion control or teleoperation. Specific issues include manipulability index\cite{torabi2018manipulability}, singularity avoiding trajectory planning\cite{marinho2016using}, and
inverse kinematics algorithms for the particular requirements of surgical teleoperation\cite{yang2016kinematic}.

\subsection{Research Using Raven}
The Raven platform users form a research community applying the system as a common experimental research platform. Much of this work has focused on issues related to \emph{control, sensing, and software}. 
A challenging frontier for surgical robotic control software, drawing increasing study, is augmenting teleoperation with autonomous functions\cite{dehghani2018automation,pedram2017autonomous,mckinley2016autonomous,hu2015semi, hu2017semi,BRL279}. 
Such functions may trade control authority back and forth between computer and surgeon, or may share control of different degrees of freedom
simultaneously with the surgeon\cite{varghese2017framework}.

Other work has used the Raven to study factors affecting \emph{teleoperation performance} \cite{milstein2018human,milstein2017scaling}. For example, experimental study focusing on effects of control parameters (in this case motion scale for the tool gripping axis) on a notion of human-centered transparency\cite{nisky2013analytical}.

Several groups have used Raven in experiments studying \emph{measurement or acquisition or updating of surgical skills} in robotic surgeons\cite{forestier2018surgical,despinoy2018evaluation,despinoy2016unsupervised}.  For example, \cite{forestier2018surgical} quantized robotic gestures into strings which were shown statistically able to discriminate skill level.  Another skill assessment study\cite{despinoy2018evaluation} compared performance and learning of trainees between a hands-on user interface device and a contactless control interface based on a low cost (compared to haptic devices) depth sensing camera and hand gestures.  They concluded that such contactless sensing has utility in training applications. The metric used to assess the interfaces was based on an unsupervised gesture recognition system\cite{despinoy2016unsupervised}.




\section{Problems and Trends in Robotic Surgeries: A Raven Perspective}
The robotic surgery techniques have made impressive progress in the past two decades. Meanwhile, with the maturation of robotic surgical techniques, the research interests shift in order to expand the application of robotic surgeries and to improve surgical outcomes. In this section, we use the Raven platform as a sample to reveal the research problems that attract attention and to discuss the open problems and the trends in the community.

\begin{figure}[h]
  \centering
  \includegraphics[width=0.65\textwidth]{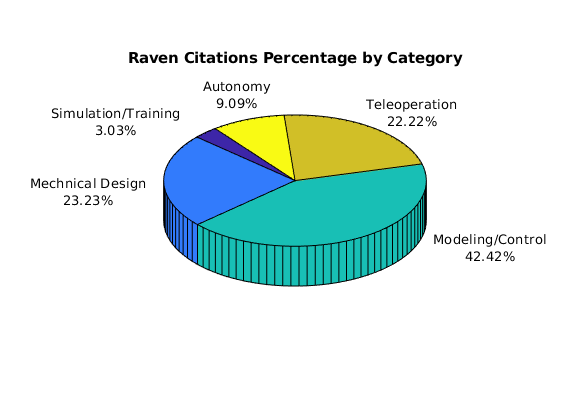}
  \caption{Raven Citations Percentage Distribution over Research Categories. It categorizes the Raven citations into five categories: 1) simulation and training, 2) mechanical design, 3)modeling and control, 4) teleoperation, 5) autonomy. The figure shows the modeling and the control is the biggest research area for the Raven related research.}
  \label{fig:categorypercentage}
\end{figure}

\begin{figure}[h]
  \centering
  \includegraphics[width=0.65\textwidth]{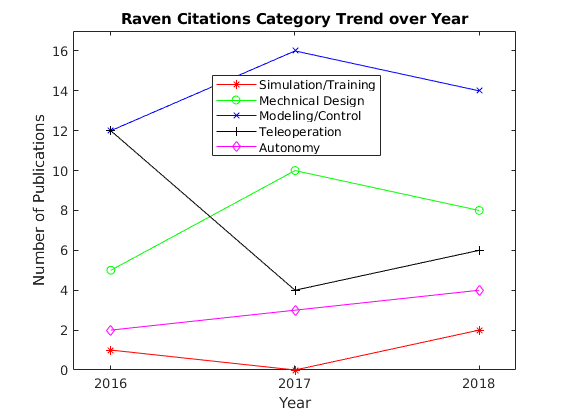}
  \caption{Categories of Raven Related Researches Trend Over Years. The mechanical design and the modeling/controlling research are dominant in raven related research. The two categories also share similar trends as the majority of the mechanical design publications also establishes the system (kinematic) models.}
  \label{fig:categoryoveryear}
\end{figure}

Raven citations reviewed above were categorized into 5 research topics:  1) simulation and training, 2) mechanical design, 3)modeling and control, 4) teleoperation, and 5) autonomy (Fig. \ref{fig:categorypercentage}). From the figure, it is clear that the modeling and control papers and the mechanical design papers are the two largest group of studies. Most of the mechanical design papers modeled the system so such papers are considered to belong to both of the two categories.  We also visualized the number of citations in each of the 5 categories with respect to years (Fig. \ref{fig:categoryoveryear}).

Under the five categories, we further divide the citations into 10 sub-categories, as shown in Table \ref{table:subcategory}, and show the numbers of citations in each of the subcategories.  In the rest of this section, we analyze the publications in the 10 sub-categories to reveal the challenge problems and the trend in the research field. 

\begin{table}[htbp]
  \centering 
  \small
  \setlength\tabcolsep{2pt}
  \resizebox{\columnwidth}{!}{
    \begin{tabular}{c | c c | c c | c c | c c c | c} 
    & & &\multicolumn{2}{c|}{Mechanical Design}  &\multicolumn{2}{c|}{Modeling \& Control} &\multicolumn{3}{c|}{Teleoperation} &Autonomy\\
    Year  &Simulation &Training &Rigid &Soft &Modeling &Learning &Master  &System &Communication &\\
    \hline 
    2016     &1     &0     &5     &0    &12     &0     &3     &3     &6     &2\\
    2017     &0     &0    &10     &0    &15     &1     &2     &1     &1     &3\\
    2018     &0     &2     &7     &1    &14     &0     &4     &1     &1     &4\\
    \end{tabular}
    }
  \caption{Publications Categorized by Topics over Years.}
  \label{table:subcategory}
\end{table}
\subsection{Modeling and Control}
Reflecting researchers' interests, most of the Raven citations discussed the modeling and the control of surgical robots\cite{niu2018improved,TII18Random,dehghani2018automation,diodato2018soft,despinoy2018evaluation,singh2018modular,JNSB17Region,yu2018pid,niu2018dimensional,xu2018kinematic,choi2018integrated,ICRA18RNNPlan,bazman2018dexterous,ICSM19Collaboration,bergelestoward,kim2017s,JNSB17AnatomicalRegion,varghese2017framework,hwang2017single,choi2017advanced,IROS17RNNAdaptiveness,pedram2017autonomous,kim2017development,RAL17GPR,megaro2017designing,anooshahpour2017modeling,ai2017design,niu2017multi,choi2017tension,choi2017integrated,IROS18RNNSoft,choi2017advanced,liu2017development,despinoy2016unsupervised,bauzano2016collaborative,hannaford2016simulation,kim2016development,beyl2016time,haghighipanah2016unscented,ai2016control,ICRA16DynamicModel,marinho2016using,ICRA16Hysteresis,bhardwaj2016preoperative,yang2016kinematic,haghighipanah2015improving}.  Most of the mechanical design publications also describe the corresponding system models and pointed out the suitable control methods.

Many of publications in this category focus on remaining challenges in modeling and control. For example, to model the control errors introduced by the driven tendons a Dahl friction model was proposed to predict cable tension for parallel robots in \cite{choi2017tension}. Some efforts were made from the hardware perspective such as the new surgical robot design and the corresponding model described in \cite{niu2018dimensional}.

There are also papers on the new challenges in the robotic surgeries. For example, a new surgical robot that delivers improved dexterity in paediatric congenital esophageal atresia surgeries\cite{kim2017development}, and a control architecture that addresses the communication, the obstacle avoidance problems in surgeon/robot collaboration\cite{bauzano2016collaborative}.

\subsection{Mechanical Design}
\label{sec:precision}
While the broader robotic community focuses on the robotic learning and the robotic vision problems, many of the efforts in the robotic surgery community were made in the area of mechanical design\cite{niu2018improved,diodato2018soft,despinoy2018evaluation,singh2018modular,niu2018dimensional,lee2018master,xu2018kinematic,bazman2018dexterous,nisar2017design,kim2017s,hwang2017single,kim2017development,megaro2017designing,anooshahpour2017modeling,ai2017design,niu2017multi,wang2017design,zhou2017development,lee2016laparoscopic,kim2016development,ai2016control,cubrich2016four,tzemanaki2016design}. This is partially because the surgical robots are quickly expanded into new surgical disciplines, in which new designs are required to meet different requirements, for example, a percutaneous surgical robot\cite{zhou2017development}. More efforts are made to improve the dexterity, the stability and the control precision of surgical robots. For example, a new surgical instrument, featuring polymer based force sensors integrated into the instrument wrist and jaws \cite{lee2016laparoscopic}, and a novel low-cost contactless optical sensor, designed to decrease the device costs and the human resource costs on training the operation of teleoperated surgical robotic systems \cite{despinoy2018evaluation}. 

\subsection{Simulation and Evaluation}
Simulation is a significant topic in robotic surgery as teleoperated robotic surgeries become more common and drive the need for a cost-efficient way to improve surgeons' skills in operating surgical robots. However, there is only one Raven citation that studies the simulation\cite{bhardwaj2016preoperative}, and only two raven citations are about evaluation\cite{forestier2018surgical,liang2018motion}.

\subsection{Teleoperated Robotic Surgeries}
As teleoperation is still the dominant way to control surgical robots, there are many Raven citations in this topic
\cite{milstein2018human,torabi2018manipulability,munteanu2018cyber,despinoy2018evaluation,lee2018master,bergelestoward,reichenbach2017telesurgery,wang2017design,frank2017ros,matsunaga2017motion,alemzadeh2016adverse,despinoy2016unsupervised,wei2016rt,alemzadeh2016targeted,lin2016safety,marinho2016using,preda2016cognitive,niu2016intuitive,tzemanaki2016design,daarchitecture,sundarapandian2016novel,yang2016kinematic}. Whilemuch research focuses on the classical teleoperation problems, such as the system architecture, the master controller and the communication problems, etc., we do see some new research problems attracting attention. 
A very important issue in a surgical robotic system is security from online adversaries.  \cite{alemzadeh2016adverse,alemzadeh2016targeted,lin2016safety}
studied cybersecurity issues {\it specific to telesurgery systems} using the Raven system.

\subsection{Autonomous Robotic Surgeries}
In a major contrast with the broader robotics community, there is only limited research on autonomy in robotic surgeries \cite{TII18Random,IROS18RNNSoft,ICRA18RNNPlan,dehghani2018automation,TII18Cooperative,pedram2017autonomous,TNNLS17Cooperative,matsunaga2017motion,JNSB17Atlas,mckinley2016autonomous,preda2016cognitive,harbison2016objective}.
Among these works, many focus on autonomizing surgical tasks, such as needle insertion \cite{dehghani2018automation} and suturing \cite{pedram2017autonomous}. There are also some works on motion planning, either motion pattern planning\cite{TII18Random,IROS18RNNSoft}, or the planning of the motion trajectories \cite{matsunaga2017motion,ICRA18RNNPlan}. 

\section{Discussion and Conclusion}
In the past two decades, surgical robotics has made notable progress and attract more roboticists worldwide. The Raven platforms, designed to enable various exploratory research through the open source software and the flexible hardware interfaces, have been serving the research community for more than 10 years. This paper used the Raven citations as a case study to review popular research problems and discusses the trends in robotic surgery study. In the literature review, we do notice some interesting phenomenon.

The surgical robot mechanical design and modeling is still the most popular research topic, according to the total number of Raven citations. More and more novel designs are proposed and developed to increase the robotic dexterity, the manipulability, the reliability, and to extend the application to new surgical procedures and address new challenges.

In contrast to the broader robotics field, machine learning, especially the deep learning, is not as popular in the surgical robotic research community. This may be due to: 1) the surgical data are often expensive so it is challenging to collect a big amount of data for training deep neural networks\cite{JNSB18RelativeMotion,JNSB17AutomatedAssess}, 2) the known challenging problems, such as environmental perception and dynamic planning, are not solved and can not reach desired reliability in the robotic surgery context\cite{NCAA18STMVO,ICRA19Segmentation,TII14IPJC,thananjeyan2017multilateral}, 3) the focus of the research community still lies on designing new robots, rather than improving robots' performance based on learning algorithms, 4) comparing with deep learning, the classical modeling methods are easier to interpolate and the performances are easier to predict, thus it is easier to predict the robot reliability.

The soft robot research in robotic surgeries is also not as popular as what we noticed in the general robotics. This may be because the Raven platforms are rigid robots and it is not straightforward to apply the soft robots on the Raven platforms or compare soft robots with the Raven.

\section*{Acknowledgment}
This work was partially supported by NSF grant IIS-1637444 and NIH grant 5R21EB016122-02.

\balance
 \newcommand{\noop}[1]{}


\begin{thebibliography}{10}

\bibitem{ravenI}
M.~J. Lum, D.~C. Friedman, G.~Sankaranarayanan, H.~King, K.~Fodero,
  R.~Leuschke, B.~Hannaford, J.~Rosen, and M.~N. Sinanan, ``The raven: Design
  and validation of a telesurgery system,'' {\em The International Journal of
  Robotics Research}, vol.~28, no.~9, pp.~1183--1197, 2009.

\bibitem{ravenII}
B.~Hannaford, J.~Rosen, D.~W. Friedman, H.~King, P.~Roan, L.~Cheng, D.~Glozman,
  J.~Ma, S.~N. Kosari, and L.~White, ``Raven-ii: an open platform for surgical
  robotics research,'' {\em IEEE Transactions on Biomedical Engineering},
  vol.~60, no.~4, pp.~954--959, 2013.

\bibitem{ravengit}
B.~Lab., ``{Raven II Software}.'' \url{https://github.com/uw-biorobotics},
  2018.

\bibitem{rosen2011surgical}
J.~Rosen, B.~Hannaford, and R.~M. Satava, {\em Surgical robotics: systems
  applications and visions}.
\newblock Springer Science \& Business Media, 2011.

\bibitem{dvrk}
P.~Kazanzides, Z.~Chen, A.~Deguet, G.~S. Fischer, R.~H. Taylor, and S.~P.
  DiMaio, ``An open-source research kit for the da vinci{\textregistered}
  surgical system,'' in {\em Robotics and Automation (ICRA), 2014 IEEE
  International Conference on}, pp.~6434--6439, IEEE, 2014.

\bibitem{kim2017development}
M.~Kim, C.~Lee, N.~Hong, Y.~J. Kim, and S.~Kim, ``Development of stereo
  endoscope system with its innovative master interface for continuous surgical
  operation,'' {\em Biomedical engineering online}, vol.~16, no.~1, p.~81,
  2017.

\bibitem{ai2017design}
Y.~Ai, B.~Pan, Y.~Fu, and S.~Wang, ``Design of a novel robotic system for
  minimally invasive surgery,'' {\em Industrial Robot: An International
  Journal}, vol.~44, no.~3, pp.~288--298, 2017.

\bibitem{reichenbach2017telesurgery}
M.~Reichenbach, T.~Frederick, L.~Cubrich, W.~Bircher, N.~Bills, M.~Morien,
  S.~Farritor, and D.~Oleynikov, ``Telesurgery with miniature robots to
  leverage surgical expertise in distributed expeditionary environments,'' {\em
  Military medicine}, vol.~182, no.~suppl\_1, pp.~316--321, 2017.

\bibitem{liu2017development}
Q.~Liu, C.~Shi, B.~Zhang, C.~Wang, L.~Duan, T.~Sun, X.~Zhang, W.~Li, Z.~Wu, and
  M.~G. Fujie, ``Development of a novel paediatric surgical assist robot for
  tissue manipulation in a narrow workspace,'' {\em Assembly Automation},
  vol.~37, no.~3, pp.~335--348, 2017.

\bibitem{zhou2017development}
C.~Zhou, H.~Wu, X.~Xu, Y.~Liu, Q.~Zhu, and S.~Pan, ``Development and control of
  a robotic arm for percutaneous surgery,'' {\em Assembly Automation}, vol.~37,
  no.~3, pp.~314--321, 2017.

\bibitem{kim2016development}
M.~Kim, C.~Lee, W.~J. Park, Y.~S. Suh, H.~K. Yang, H.~J. Kim, and S.~Kim, ``A
  development of assistant surgical robot system based on
  surgical-operation-by-wire and hands-on-throttle-and-stick,'' {\em Biomedical
  engineering online}, vol.~15, no.~1, p.~58, 2016.

\bibitem{marinho2016using}
M.~M. Marinho, M.~C. Bernardes, and A.~P. Bo, ``Using general-purpose
  serial-link manipulators for laparoscopic surgery with moving remote center
  of motion,'' {\em Journal of Medical Robotics Research}, vol.~1, no.~04,
  p.~1650007, 2016.

\bibitem{niu2018improved}
G.~Niu, B.~Pan, F.~Zhang, H.~Feng, and Y.~Fu, ``Improved surgical instruments
  without coupled motion used in minimally invasive surgery,'' {\em The
  International Journal of Medical Robotics and Computer Assisted Surgery},
  p.~e1942, 2018.

\bibitem{diodato2018soft}
A.~Diodato, M.~Brancadoro, G.~De~Rossi, H.~Abidi, D.~Dall’Alba, R.~Muradore,
  G.~Ciuti, P.~Fiorini, A.~Menciassi, and M.~Cianchetti, ``Soft robotic
  manipulator for improving dexterity in minimally invasive surgery,'' {\em
  Surgical innovation}, vol.~25, no.~1, pp.~69--76, 2018.

\bibitem{niu2018dimensional}
G.~Niu, B.~Pan, F.~Zhang, H.~Feng, W.~Gao, and Y.~Fu, ``Dimensional synthesis
  and concept design of a novel minimally invasive surgical robot,'' {\em
  Robotica}, vol.~36, no.~5, pp.~715--737, 2018.

\bibitem{xu2018kinematic}
W.~Xu, Y.~Wang, S.~Jiang, J.~Yao, and B.~Chen, ``Kinematic analysis of a newly
  designed cable-driven manipulator,'' {\em Transactions of the Canadian
  Society for Mechanical Engineering}, vol.~42, no.~2, pp.~125--135, 2018.

\bibitem{bazman2018dexterous}
M.~Bazman, N.~Yilmaz, and U.~Tumerdem, ``Dexterous and back-drivable parallel
  robotic forceps wrist for robotic surgery,'' in {\em Advanced Motion Control
  (AMC), 2018 IEEE 15th International Workshop on}, pp.~153--159, IEEE, 2018.

\bibitem{nisar2017design}
S.~Nisar, T.~Endo, and F.~Matsuno, ``Design and kinematic optimization of a two
  degrees-of-freedom planar remote center of motion mechanism for minimally
  invasive surgery manipulators,'' {\em Journal of Mechanisms and Robotics},
  vol.~9, no.~3, p.~031013, 2017.

\bibitem{kim2017s}
U.~Kim, D.-H. Lee, Y.~B. Kim, D.-Y. Seok, J.~So, and H.~R. Choi, ``S-surge:
  Novel portable surgical robot with multiaxis force-sensing capability for
  minimally invasive surgery,'' {\em IEEE/ASME Transactions on Mechatronics},
  vol.~22, no.~4, pp.~1717--1727, 2017.

\bibitem{hwang2017single}
M.~Hwang, U.-J. Yang, D.~Kong, D.~G. Chung, J.-g. Lim, D.-H. Lee, D.~H. Kim,
  D.~Shin, T.~Jang, J.-W. Kim, {\em et~al.}, ``A single port surgical robot
  system with novel elbow joint mechanism for high force transmission,'' {\em
  The International Journal of Medical Robotics and Computer Assisted Surgery},
  vol.~13, no.~4, p.~e1808, 2017.

\bibitem{rosen2017roboscope}
J.~Rosen, L.~N. Sekhar, D.~Glozman, M.~Miyasaka, J.~Dosher, B.~Dellon, K.~S.
  Moe, A.~Kim, L.~J. Kim, T.~Lendvay, {\em et~al.}, ``Roboscope: A flexible and
  bendable surgical robot for single portal minimally invasive surgery,'' in
  {\em Robotics and Automation (ICRA), 2017 IEEE International Conference on},
  pp.~2364--2370, IEEE, 2017.

\bibitem{cubrich2016four}
L.~Cubrich, M.~A. Reichenbach, J.~D. Carlson, A.~Pracht, B.~Terry,
  D.~Oleynikov, and S.~Farritor, ``A four-dof laparo-endoscopic single site
  platform for rapidly-developing next-generation surgical robotics,'' {\em
  Journal of Medical Robotics Research}, vol.~1, no.~04, p.~1650006, 2016.

\bibitem{niu2017multi}
G.-j. Niu, B.~Pan, F.-h. Zhang, H.-b. Feng, and Y.-l. Fu, ``Multi-optimization
  of a spherical mechanism for minimally invasive surgery,'' {\em Journal of
  Central South University}, vol.~24, no.~6, pp.~1406--1417, 2017.

\bibitem{torabi2018manipulability}
A.~Torabi, M.~Khadem, K.~Zareinia, G.~R. Sutherland, and M.~Tavakoli,
  ``Manipulability of teleoperated surgical robots with application in design
  of master/slave manipulators,'' in {\em Medical Robotics (ISMR), 2018
  International Symposium on}, pp.~1--6, IEEE, 2018.

\bibitem{singh2018modular}
S.~Singh, A.~Singla, and E.~Singla, ``Modular manipulators for cluttered
  environments: A task-based configuration design approach,'' {\em Journal of
  Mechanisms and Robotics}, vol.~10, no.~5, p.~051010, 2018.

\bibitem{yang2016kinematic}
D.~Yang, L.~Wang, and Y.~Li, ``Kinematic analysis and simulation of a misr
  system using bimanual manipulator,'' in {\em Robotics and Biomimetics
  (ROBIO), 2016 IEEE International Conference on}, pp.~271--276, IEEE, 2016.

\bibitem{dehghani2018automation}
H.~Dehghani, S.~Farritor, D.~Oleynikov, and B.~Terry, ``Automation of suturing
  path generation for da vinci-like surgical robotic systems,'' in {\em 2018
  Design of Medical Devices Conference}, pp.~V001T07A008--V001T07A008, American
  Society of Mechanical Engineers, 2018.

\bibitem{pedram2017autonomous}
S.~A. Pedram, P.~Ferguson, J.~Ma, E.~Dutson, and J.~Rosen, ``Autonomous
  suturing via surgical robot: An algorithm for optimal selection of needle
  diameter, shape, and path,'' in {\em Robotics and Automation (ICRA), 2017
  IEEE International Conference on}, pp.~2391--2398, IEEE, 2017.

\bibitem{mckinley2016autonomous}
S.~McKinley, A.~Garg, S.~Sen, D.~V. Gealy, J.~P. McKinley, Y.~Jen, and
  K.~Goldberg, ``Autonomous multilateral surgical tumor resection with
  interchangeable instrument mounts and fluid injection device,'' in {\em 2016
  IEEE International Conference on Robotics and Automation (ICRA)}, 2016.

\bibitem{hu2015semi}
D.~Hu, Y.~Gong, B.~Hannaford, and E.~J. Seibel, ``Semi-autonomous simulated
  brain tumor ablation with ravenii surgical robot using behavior tree,'' in
  {\em 2015 IEEE International Conference on Robotics and Automation (ICRA)},
  pp.~3868--3875, IEEE, 2015.

\bibitem{hu2017semi}
D.~Hu, Y.~Gong, E.~J. Seibel, L.~N. Sekhar, and B.~Hannaford, ``Semi-autonomous
  image-guided brain tumour resection using an integrated robotic system: A
  bench-top study,'' {\em The International Journal of Medical Robotics and
  Computer Assisted Surgery}, 2017.

\bibitem{BRL279}
D.~Hu, Y.~Gong, B.~Hannaford, and E.~J. Seibel, ``Path planning for
  semi-automated simulated robotic neurosurgery,'' in {\em 2015 IEEE/RSJ
  International Conference on Intelligent Robots and Systems (IROS)}, September
  2015.

\bibitem{varghese2017framework}
R.~J. Varghese, P.~Berthet-Rayne, P.~Giataganas, V.~Vitiello, and G.-Z. Yang,
  ``A framework for sensorless and autonomous probe-tissue contact management
  in robotic endomicroscopic scanning,'' in {\em Robotics and Automation
  (ICRA), 2017 IEEE International Conference on}, pp.~1738--1745, IEEE, 2017.

\bibitem{milstein2018human}
A.~Milstein, T.~Ganel, S.~Berman, and I.~Nisky, ``Human-centered transparency
  of grasping via a robot-assisted minimally invasive surgery system,'' {\em
  IEEE Transactions on Human-Machine Systems}, vol.~48, no.~4, pp.~349--358,
  2018.

\bibitem{milstein2017scaling}
A.~Milstein, T.~Ganel, S.~Berman, and I.~Nisky, ``The scaling of the gripper
  affects the action and perception in teleoperated grasping via a
  robot-assisted minimally invasive surgery system,'' {\em arXiv preprint
  arXiv:1710.05319}, 2017.

\bibitem{nisky2013analytical}
I.~Nisky, F.~A. Mussa-Ivaldi, and A.~Karniel, ``Analytical study of perceptual
  and motor transparency in bilateral teleoperation,'' {\em IEEE Transactions
  on Human-Machine Systems}, vol.~43, no.~6, pp.~570--582, 2013.

\bibitem{forestier2018surgical}
G.~Forestier, F.~Petitjean, P.~Senin, F.~Despinoy, A.~Huaulm{\'e}, H.~I. Fawaz,
  J.~Weber, L.~Idoumghar, P.-A. Muller, and P.~Jannin, ``Surgical motion
  analysis using discriminative interpretable patterns,'' {\em Artificial
  intelligence in medicine}, 2018.

\bibitem{despinoy2018evaluation}
F.~Despinoy, N.~Zemiti, G.~Forestier, A.~S{\'a}nchez, P.~Jannin, and
  P.~Poignet, ``Evaluation of contactless human--machine interface for robotic
  surgical training,'' {\em International journal of computer assisted
  radiology and surgery}, vol.~13, no.~1, pp.~13--24, 2018.

\bibitem{despinoy2016unsupervised}
F.~Despinoy, D.~Bouget, G.~Forestier, C.~Penet, N.~Zemiti, P.~Poignet, and
  P.~Jannin, ``Unsupervised trajectory segmentation for surgical gesture
  recognition in robotic training,'' {\em IEEE Transactions on Biomedical
  Engineering}, vol.~63, no.~6, pp.~1280--1291, 2016.

\bibitem{TII18Random}
Y.~Li, S.~Li, and B.~Hannaford, ``A model based recurrent neural network with
  randomness for efficient control with applications,'' {\em IEEE Transactions
  on Industrial Informatics}, 2018.

\bibitem{JNSB17Region}
R.~A. Harbison, A.~M. Berens, Y.~Li, R.~A. Bly, B.~Hannaford, and K.~S. Moe,
  ``Region-specific objective signatures of endoscopic surgical instrument
  motion: A cadaveric exploratory analysis,'' {\em Journal of Neurological
  Surgery Part B: Skull Base}, vol.~78, no.~01, pp.~099--104, 2017.

\bibitem{yu2018pid}
W.~Yu, {\em PID Control with Intelligent Compensation for Exoskeleton Robots}.
\newblock Academic Press, 2018.

\bibitem{choi2018integrated}
S.-H. Choi and K.-S. Park, ``Integrated and nonlinear dynamic model of a
  polymer cable for low-speed cable-driven parallel robots,'' {\em Microsystem
  Technologies}, pp.~1--11, 2018.

\bibitem{ICRA18RNNPlan}
Y.~Li, S.~Li, and B.~Hannaford, ``A novel recurrent neural network control
  scheme for improving redundant manipulator motion planning completeness,'' in
  {\em Robotics and Automation (ICRA), 2018 IEEE International Conference on},
  p.~1~6, IEEE, 2018.

\bibitem{ICSM19Collaboration}
Y.~Li, ``Trends in control and decision-making for human-robot collaboration
  systems [bookshelf],'' {\em IEEE Control Systems Magazine}, vol.~39,
  pp.~101--103, April 2019.

\bibitem{bergelestoward}
C.~Bergeles, ``Toward intracorporeally navigated untethered microsurgeons,''

\bibitem{JNSB17AnatomicalRegion}
Y.~Li, R.~A. Bly, R.~A. Harbison, I.~M. Humphreys, M.~E. Whipple, B.~Hannaford,
  and K.~S. Moe, ``Anatomical region segmentation for objective surgical skill
  assessment with operating room motion data,'' {\em Journal of Neurological
  Surgery Part B: Skull Base}, vol.~369, no.~15, pp.~1434--1442, 2017.

\bibitem{choi2017advanced}
S.-H. Choi and K.-S. Park, ``Advanced numerical modeling of nonlinear elastic
  cable with recovery characteristics,'' in {\em ASME 2017 Conference on
  Information Storage and Processing Systems collocated with the ASME 2017
  Conference on Information Storage and Processing Systems},
  pp.~V001T07A009--V001T07A009, American Society of Mechanical Engineers, 2017.

\bibitem{IROS17RNNAdaptiveness}
Y.~Li, S.~Li, M.~Miyasaka, A.~Lewis, and B.~Hannaford, ``Improving control
  precision and motion adaptiveness for surgical robot with recurrent neural
  network,'' in {\em Intelligent Robots and Systems (IROS), 2017 IEEE/RSJ
  International Conference on}, pp.~1--6, IEEE, 2017.

\bibitem{RAL17GPR}
Y.~Li and B.~Hannaford, ``Gaussian process regression for sensorless grip force
  estimation of cable-driven elongated surgical instruments,'' {\em IEEE
  Robotics and Automation Letters}, vol.~2, no.~3, pp.~1312--1319, 2017.

\bibitem{megaro2017designing}
V.~Megaro, E.~Knoop, A.~Spielberg, D.~I. Levin, W.~Matusik, M.~Gross,
  B.~Thomaszewski, and M.~B{\"a}cher, ``Designing cable-driven actuation
  networks for kinematic chains and trees,'' in {\em Proceedings of the ACM
  SIGGRAPH/Eurographics Symposium on Computer Animation}, p.~15, ACM, 2017.

\bibitem{anooshahpour2017modeling}
F.~Anooshahpour, P.~Yadmellat, I.~G. Polushin, and R.~V. Patel, ``Modeling of
  tendon-pulley transmission systems with application to surgical robots: A
  preliminary experimental validation,'' in {\em Advanced Intelligent
  Mechatronics (AIM), 2017 IEEE International Conference on}, pp.~761--766,
  IEEE, 2017.

\bibitem{choi2017tension}
S.-H. Choi, J.-O. Park, and K.-S. Park, ``Tension analysis of a
  6-degree-of-freedom cable-driven parallel robot considering dynamic pulley
  bearing friction,'' {\em Advances in Mechanical Engineering}, vol.~9, no.~8,
  p.~1687814017714981, 2017.

\bibitem{choi2017integrated}
S.-H. Choi and K.-S. Park, ``The integrated elasto-plastic cable modeling for
  cable driven parallel robots (cdprs),'' in {\em Control, Automation and
  Systems (ICCAS), 2017 17th International Conference on}, pp.~420--422, IEEE,
  2017.

\bibitem{IROS18RNNSoft}
Y.~Li and B.~Hannaford, ``Soft-obstacle avoidance for redundant manipulators
  with recurrent neural network,'' in {\em Intelligent Robots and Systems
  (IROS), 2018 IEEE/RSJ International Conference on}, pp.~1--6, IEEE, 2018.

\bibitem{bauzano2016collaborative}
E.~Bauzano, B.~Estebanez, I.~Garcia-Morales, and V.~F. Mu{\~n}oz,
  ``Collaborative human--robot system for hals suture procedures,'' {\em IEEE
  Systems Journal}, vol.~10, no.~3, pp.~957--966, 2016.

\bibitem{hannaford2016simulation}
B.~Hannaford, D.~Hu, D.~Zhang, and Y.~Li, ``Simulation results on selector
  adaptation in behavior trees,'' 2016.

\bibitem{beyl2016time}
T.~Beyl, P.~Nicolai, M.~D. Comparetti, J.~Raczkowsky, E.~De~Momi, and
  H.~W{\"o}rn, ``Time-of-flight-assisted kinect camera-based people detection
  for intuitive human robot cooperation in the surgical operating room,'' {\em
  International journal of computer assisted radiology and surgery}, vol.~11,
  no.~7, pp.~1329--1345, 2016.

\bibitem{haghighipanah2016unscented}
M.~Haghighipanah, M.~Miyasaka, Y.~Li, and B.~Hannaford, ``Unscented kalman
  filter and 3d vision to improve cable driven surgical robot joint angle
  estimation,'' in {\em Robotics and Automation (ICRA), 2016 IEEE International
  Conference on}, pp.~4135--4142, IEEE, 2016.

\bibitem{ai2016control}
Y.~Ai, B.~Pan, Y.~Fu, and S.~Wang, ``Control system design for a novel
  minimally invasive surgical robot,'' {\em Computer Assisted Surgery},
  vol.~21, no.~sup1, pp.~45--53, 2016.

\bibitem{ICRA16DynamicModel}
Y.~Li, M.~Miyasaka, M.~Haghighipanah, L.~Cheng, and B.~Hannaford, ``Dynamic
  modeling of cable driven elongated surgical instruments for sensorless grip
  force estimation,'' in {\em Robotics and Automation (ICRA), 2016 IEEE
  International Conference on}, pp.~4128--4134, IEEE, 2016.

\bibitem{ICRA16Hysteresis}
M.~Miyasaka, M.~Haghighipanah, Y.~Li, and B.~Hannaford, ``Hysteresis model of
  longitudinally loaded cable for cable driven robots and identification of the
  parameters,'' in {\em Robotics and Automation (ICRA), 2016 IEEE International
  Conference on}, pp.~4051--4057, IEEE, 2016.

\bibitem{bhardwaj2016preoperative}
A.~Bhardwaj, A.~Jain, and V.~Agarwal, ``Preoperative planning simulator with
  haptic feedback for raven-ii surgical robotics platform,'' in {\em Computing
  for Sustainable Global Development (INDIACom), 2016 3rd International
  Conference on}, pp.~2443--2448, IEEE, 2016.

\bibitem{haghighipanah2015improving}
M.~Haghighipanah, Y.~Li, M.~Miyasaka, and B.~Hannaford, ``Improving position
  precision of a servo-controlled elastic cable driven surgical robot using
  unscented kalman filter,'' in {\em Intelligent Robots and Systems (IROS),
  2015 IEEE/RSJ International Conference on}, pp.~2030--2036, IEEE, 2015.

\bibitem{lee2018master}
H.~Lee, B.~Cheon, M.~Hwang, D.~Kang, and D.-S. Kwon, ``A master manipulator
  with a remote-center-of-motion kinematic structure for a minimally invasive
  robotic surgical system,'' {\em The International Journal of Medical Robotics
  and Computer Assisted Surgery}, vol.~14, no.~1, p.~e1865, 2018.

\bibitem{wang2017design}
T.~Wang, B.~Pan, Y.~Fu, S.~Wang, and Y.~Ai, ``Design of a new haptic device and
  experiments in minimally invasive surgical robot,'' {\em Computer Assisted
  Surgery}, vol.~22, no.~sup1, pp.~240--250, 2017.

\bibitem{lee2016laparoscopic}
D.-H. Lee, U.~Kim, T.~Gulrez, W.~J. Yoon, B.~Hannaford, and H.~R. Choi, ``A
  laparoscopic grasping tool with force sensing capability,'' {\em IEEE/ASME
  Transactions on Mechatronics}, vol.~21, no.~1, pp.~130--141, 2016.

\bibitem{tzemanaki2016design}
A.~Tzemanaki, L.~Fracczak, D.~Gillatt, A.~Koupparis, C.~Melhuish, R.~Persad,
  E.~Rowe, A.~G. Pipe, and S.~Dogramadzi, ``Design of a multi-dof cable-driven
  mechanism of a miniature serial manipulator for robot-assisted minimally
  invasive surgery,'' in {\em Biomedical Robotics and Biomechatronics (BioRob),
  2016 6th IEEE International Conference on}, pp.~55--60, IEEE, 2016.

\bibitem{liang2018motion}
K.~Liang, Y.~Xing, J.~Li, S.~Wang, A.~Li, and J.~Li, ``Motion control skill
  assessment based on kinematic analysis of robotic end-effector movements,''
  {\em The International Journal of Medical Robotics and Computer Assisted
  Surgery}, vol.~14, no.~1, p.~e1845, 2018.

\bibitem{munteanu2018cyber}
A.~Munteanu, R.~Muradore, M.~Merro, and P.~Fiorini, ``On cyber-physical attacks
  in bilateral teleoperation systems: An experimental analysis,'' in {\em 2018
  IEEE Industrial Cyber-Physical Systems (ICPS)}, pp.~159--166, IEEE, 2018.

\bibitem{frank2017ros}
T.~Frank, A.~Krieger, S.~Leonard, N.~A. Patel, and J.~Tokuda,
  ``Ros-igtl-bridge: an open network interface for image-guided therapy using
  the ros environment,'' {\em International journal of computer assisted
  radiology and surgery}, vol.~12, no.~8, pp.~1451--1460, 2017.

\bibitem{matsunaga2017motion}
T.~Matsunaga, T.~Okano, X.~Sun, T.~Mizoguchi, and K.~Ohnishi, ``Motion
  reproduction system using multi dof haptic forceps robots for ligation
  task,'' in {\em Industrial Electronics Society, IECON 2017-43rd Annual
  Conference of the IEEE}, pp.~2888--2893, IEEE, 2017.

\bibitem{alemzadeh2016adverse}
H.~Alemzadeh, J.~Raman, N.~Leveson, Z.~Kalbarczyk, and R.~K. Iyer, ``Adverse
  events in robotic surgery: a retrospective study of 14 years of fda data,''
  {\em PloS one}, vol.~11, no.~4, p.~e0151470, 2016.

\bibitem{wei2016rt}
H.~Wei, Z.~Shao, Z.~Huang, R.~Chen, Y.~Guan, J.~Tan, and Z.~Shao, ``Rt-ros: A
  real-time ros architecture on multi-core processors,'' {\em Future Generation
  Computer Systems}, vol.~56, pp.~171--178, 2016.

\bibitem{alemzadeh2016targeted}
H.~Alemzadeh, D.~Chen, X.~Li, T.~Kesavadas, Z.~T. Kalbarczyk, and R.~K. Iyer,
  ``Targeted attacks on teleoperated surgical robots: Dynamic model-based
  detection and mitigation,'' in {\em Dependable Systems and Networks (DSN),
  2016 46th Annual IEEE/IFIP International Conference on}, pp.~395--406, IEEE,
  2016.

\bibitem{lin2016safety}
H.~Lin, H.~Alemzadeh, D.~Chen, Z.~Kalbarczyk, and R.~K. Iyer, ``Safety-critical
  cyber-physical attacks: Analysis, detection, and mitigation,'' in {\em
  Proceedings of the Symposium and Bootcamp on the Science of Security},
  pp.~82--89, ACM, 2016.

\bibitem{preda2016cognitive}
N.~Preda, F.~Ferraguti, G.~De~Rossi, C.~Secchi, R.~Muradore, P.~Fiorini, and
  M.~Bonf{\`e}, ``A cognitive robot control architecture for autonomous
  execution of surgical tasks,'' {\em Journal of Medical Robotics Research},
  vol.~1, no.~04, p.~1650008, 2016.

\bibitem{niu2016intuitive}
G.~Niu, B.~Pan, Y.~Ai, and Y.~Fu, ``Intuitive control algorithm of a novel
  minimally invasive surgical robot,'' {\em Computer Assisted Surgery},
  vol.~21, no.~sup1, pp.~92--101, 2016.

\bibitem{daarchitecture}
X.~Da, Q.~Cao, P.~Chen, and M.~Adachi, ``Architecture design for a mis robot
  control system,''

\bibitem{sundarapandian2016novel}
S.~Sundarapandian, S.~Ferris-Francis, H.~Michko, C.~Charters, J.~Miller, and
  N.~Prabakar, ``A novel communication architecture and control system for
  telebot: A multi-modal telepresence robot for disabled officers.,'' {\em
  International Journal of Next-Generation Computing}, vol.~7, no.~3, 2016.

\bibitem{TII18Cooperative}
L.~Jin, S.~Li, X.~Luo, Y.~Li, and B.~Qin, ``Neural dynamics for cooperative
  control of redundant robot manipulators,'' {\em IEEE Transactions on
  Industrial Informatics}, vol.~14, no.~9, pp.~3812--3821, 2018.

\bibitem{TNNLS17Cooperative}
S.~Li, J.~He, Y.~Li, and M.~U. Rafique, ``Distributed recurrent neural networks
  for cooperative control of manipulators: A game-theoretic perspective,'' {\em
  IEEE transactions on neural networks and learning systems}, vol.~28, no.~2,
  pp.~415--426, 2017.

\bibitem{JNSB17Atlas}
Y.~Li, R.~A. Harbison, R.~A. Bly, I.~M. Humphreys, B.~Hannaford, and K.~Moe,
  ``Atlas based anatomical region segmentation for minimally invasive skull
  base surgery objective motion analysis,'' in {\em Journal of Neurological
  Surgery Part B: Skull Base}, vol.~78, p.~A146, Georg Thieme Verlag KG, 2017.

\bibitem{harbison2016objective}
R.~A. Harbison, A.~Berens, Y.~Li, A.~Law, M.~Whipple, B.~Hannaford, and K.~Moe,
  ``Objective signatures of endoscopic surgical performance,'' in {\em Journal
  of Neurological Surgery Part B: Skull Base}, vol.~77, p.~A120, 2016.

\bibitem{JNSB18RelativeMotion}
Y.~Li, R.~Bly, M.~Whipple, I.~Humphreys, B.~Hannaford, and K.~Moe, ``Use
  endoscope and instrument and pathway relative motion as metric for automated
  objective surgical skill assessment in skull base and sinus surgery,'' in
  {\em Journal of Neurological Surgery Part B: Skull Base}, vol.~79, p.~A194,
  Georg Thieme Verlag KG, 2018.

\bibitem{JNSB17AutomatedAssess}
R.~A. Harbison, Y.~Li, A.~M. Berens, R.~A. Bly, B.~Hannaford, and K.~S. Moe,
  ``An automated methodology for assessing anatomy-specific instrument motion
  during endoscopic endonasal skull base surgery,'' {\em Journal of
  Neurological Surgery Part B: Skull Base}, vol.~38, no.~03, pp.~222--226,
  2017.

\bibitem{NCAA18STMVO}
Y.~Li, J.~Zhang, and S.~Li, ``{STMVO}: biologically inspired monocular visual
  odometry,'' {\em Neural Computing and Applications}, vol.~29, no.~6,
  pp.~215--225, 2018.

\bibitem{ICRA19Segmentation}
F.~Qin, Y.~Li, Y.-H. Su, D.~Xu, and B.~Hannaford, ``Surgical instrument
  segmentation for endoscopic vision with data fusion of cnn prediction and
  kinematic pose,'' in {\em Robotics and Automation (ICRA), 2019 IEEE
  International Conference on}, pp.~1--6, IEEE, 2019.

\bibitem{TII14IPJC}
Y.~Li, S.~Li, Q.~Song, H.~Liu, and M.~Q.-H. Meng, ``Fast and robust data
  association using posterior based approximate joint compatibility test,''
  {\em IEEE Transactions on Industrial Informatics}, vol.~10, no.~1,
  pp.~331--339, 2014.

\bibitem{thananjeyan2017multilateral}
B.~Thananjeyan, A.~Garg, S.~Krishnan, C.~Chen, L.~Miller, and K.~Goldberg,
  ``Multilateral surgical pattern cutting in 2d orthotropic gauze with deep
  reinforcement learning policies for tensioning,'' in {\em 2017 IEEE
  International Conference on Robotics and Automation (ICRA)}, pp.~2371--2378,
  IEEE, 2017.

\end{thebibliography}
\end{document}